\documentclass[conference]{IEEEtran}
\IEEEoverridecommandlockouts
\usepackage{cite}
\usepackage{amsmath,amssymb,amsfonts}
\usepackage{algorithmic}
\usepackage{graphicx}
\usepackage{textcomp}
\usepackage{xcolor}
\usepackage[utf8]{inputenc}
\usepackage{adjustbox}
\usepackage{multirow}

\usepackage{booktabs}

\def\BibTeX{{\rm B\kern-.05em{\sc i\kern-.025em b}\kern-.08em
    T\kern-.1667em\lower.7ex\hbox{E}\kern-.125emX}}
\usepackage{xspace}
\newcommand{\we}{\textit{BalanceGS}\xspace}

\begin{document}

\title{\we: Algorithm-System Co-design for Efficient 3D Gaussian Splatting Training on GPU}

\author{Junyi Wu$^{1*}$, Jiaming Xu$^{1,3*}$, Jinhao Li$^{1,3}$, Yongkang Zhou$^{1,3}$, Jiayi Pan$^{1,2}$, \\Xingyang Li$^{1}$, Guohao Dai$^{1,2,3\dag}$ \\
$^{1}$Shanghai Jiao Tong University, $^{2}$Infinigence-AI, $^{3}$SII \\
$^*$Equal contribution, $^{\dag}$Corresponding author: daiguohao@sjtu.edu.cn
\vspace{-16pt}}

\maketitle

\begin{abstract}
3D Gaussian Splatting (3DGS) has emerged as a promising 3D reconstruction technique. 
The traditional 3DGS training pipeline follows three sequential steps: \textit{Gaussian densification}, \textit{Gaussian projection}, and \textit{color splatting}. Despite its promising reconstruction quality, this conventional approach suffers from three critical inefficiencies:
(1) \underline{Skewed} density allocation during Gaussian densification. The adaptive densification strategy in 3DGS makes skewed Gaussian allocation across dense and sparse regions. The number of Gaussians of dense regions can be $100 \times$ that of sparse regions, leading to Gaussian redundancy. 
(2) \underline{Imbalanced} computation workload during Gaussian projection. The traditional one-to-one allocation mechanism between threads and pixels results in execution time discrepancies between threads, leading to  $\sim20\%$ latency overhead.
(3) \underline{Fragmented} memory access during color splatting. Discrete storage of colors in memory fails to take advantage of data locality with fragmented memory access, resulting in $\sim 2.0\times$ color memory access time.

To tackle the above challenges, we introduce \we, the algorithm-system co-design for efficient training in 3DGS. (1) At the algorithm level, we propose \underline{heuristic workload-sensitive} Gaussian density control to automatically balance point distributions - removing 80\% redundant Gaussians in dense regions while filling gaps in sparse areas.
(2) At the system level, we propose \underline{Similarity-based} Gaussian sampling and merging, which replaces the static one-to-one thread-pixel mapping with adaptive workload distribution - threads now dynamically process variable numbers of Gaussians based on local cluster density.
(3) At the mapping level, we propose \underline{reordering-based} memory access mapping strategy that restructures RGB storage and enables batch loading in shared memory.  

Extensive experiments demonstrate that compared with 3DGS, our approach achieves a 1.44$\times$ training speedup on a NVIDIA A100 GPU with negligible quality degradation.

\end{abstract}

\section{Introduction}

3D Gaussian Splatting (3DGS)~\cite{kerbl3Dgaussians} has emerged as a breakthrough technique for real-time 3D scene representation, utilizing spatially distributed Gaussians as rendering primitives. 
In SLAM~\cite{Matsuki:Murai:etal:CVPR2024,yan2023gs,keetha2024splatam,hhuang2024photoslam}, 3DGS enables real-time 3D mapping by incrementally building and updating Gaussian representations from streaming images. For autonomous driving~\cite{zhou2024drivinggaussian,chen2024omnire,yan2024street,Zhou_2024_CVPR}, 3DGS efficiently represents moving objects like vehicles and pedestrians. In digital avatars~\cite{rivero2024rig3dgs,stanishevskii2024implicitdeepfakeplausiblefaceswappingimplicit,liu24-GVA,zhou2024headstudio}, the technique enables high-fidelity facial animation through deformable Gaussian representations that balance quality and rendering speed.

\begin{figure*}[htbp]
    \centering
    \includegraphics[width=0.99\textwidth]{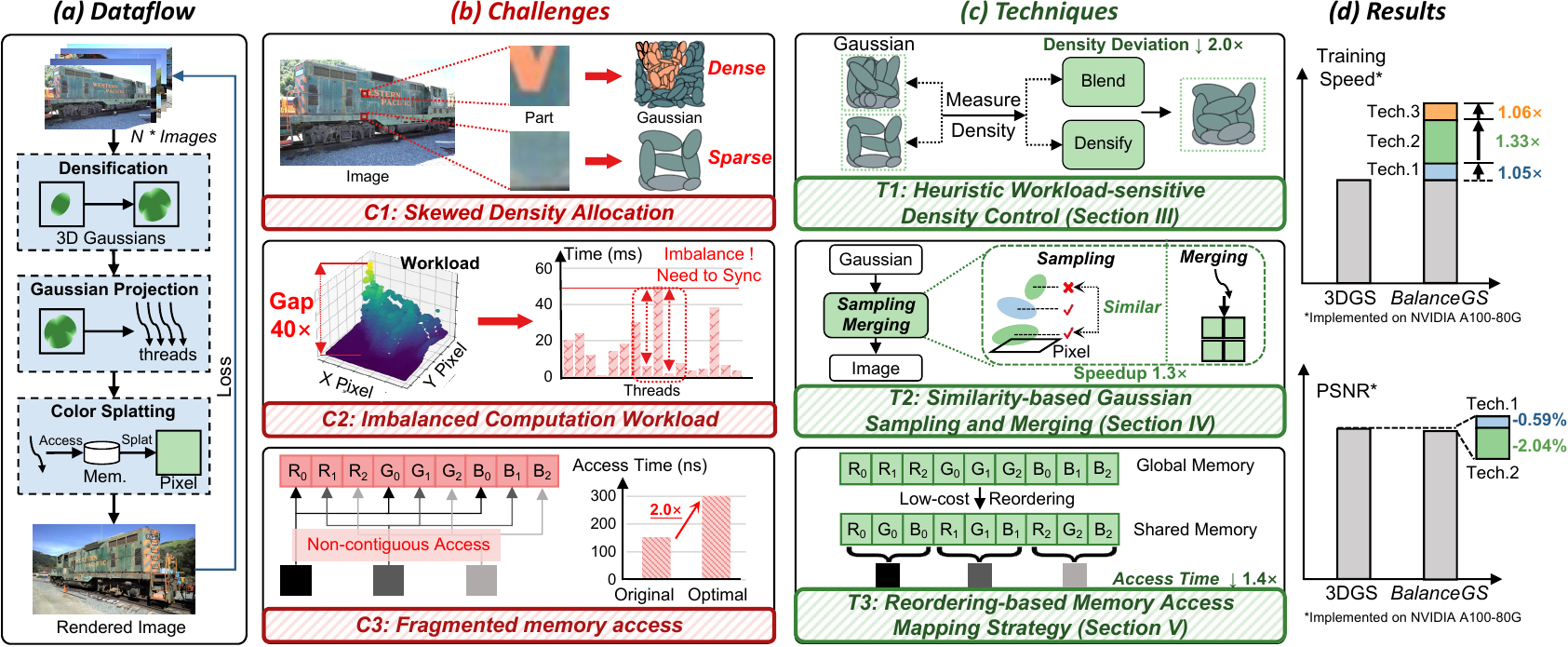}
    \vspace{-10pt}
    \caption{Overview of BalanceGS: (a) training dataflow of 3DGS. (b) three challenges, and (c) three proposed techniques}
    \vspace{-5pt}
    \label{fig:overview}
\end{figure*}

As is shown in Fig.~\ref{fig:overview}, the training pipeline of 3DGS follows three sequential steps.
\textit{(1) Gaussian densification} establishes scene representation through adaptive density control in training~\cite{kerbl3Dgaussians}, where Gaussians are strategically cloned and split in under-reconstruction and over-reconstruction regions.
\textit{(2) Gaussian projection} then transforms these 3D primitives into 2D splats for later color splatting, projecting of Gaussains to screen space.
\textit{(3) Color splatting} finally computes per-pixel appearance using spherical harmonics (SH) coefficients. 

\begin{figure}[t]
    \centering
    \includegraphics[width=0.98\linewidth]{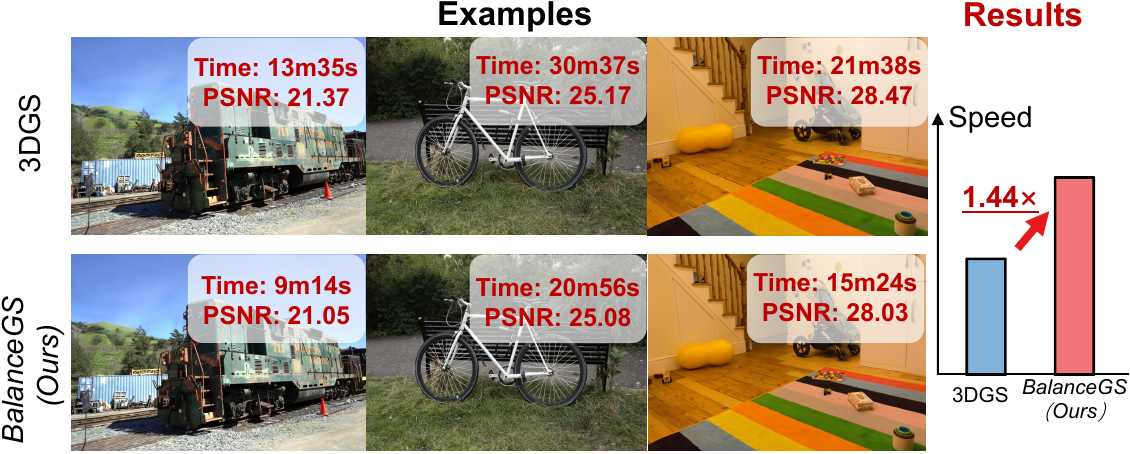}
    \vspace{-5pt}
    \caption{Example and result comparsion between  3D Gaussian Splatting and \we}
    \vspace{-15pt}
    \label{fig1}
\end{figure}

While numerous optimizations have been proposed from algorithmic~\cite{hac2024} and system-level perspectives~\cite{lee2024compact}, our analysis reveals persistent performance bottlenecks originating from hierarchical resource inefficiencies across geometry, computation and memory, as illustrated in Fig.~\ref{fig:overview}. However, the following challenges remain unsolved to
accelerate end-to-end training pipeline in Fig.~\ref{fig:overview}(a).

\textbf{\textit{Challenge-1}: Skewed density allocation during Gaussian densification.}
As illustrated in Fig.~\ref{fig:overview}-C1, imbalanced densification leads to redundant Gaussian accumulation, with density variations reaching 100$\times$ between dense and sparse areas. This skewed density allocation not only wastes computational resources but also causes quality degradation where additional Gaussians yield negative effect. Meanwhile, sparse regions suffer from insufficient coverage, illustrating rendering artifacts in low-texture areas. Current adaptive densification strategies create skewed density allocation during Gaussian densification. 

\textbf{\textit{Challenge-2}: Imbalanced computation workload during Gaussian projection.}
The standard one-to-one thread-to-pixel allocation fails to account for the highly non-uniform distribution of Gaussians across the scene. While this mapping works well for traditional rasterization where each pixel requires similar computation, in 3DGS dense regions may require processing 40× more Gaussians per pixel than sparse regions. This workload imbalance leads to severe thread underutilization, with 20\% of GPU cycles wasted on synchronization.

\textbf{\textit{Challenge-3}: Fragmented memory access during color splatting.} 
The color splatting stage requires sequential RGB retrieval for each Gaussian, but the default Structure-of-Arrays (SoA) storage ([R1...Rn], [G1...Gn], [B1...Bn]) scatters each color channel across separate memory regions. This conflict between the access pattern (sequential per-Gaussian RGB) and storage layout (channel-wise separation) forces discontinuous global memory fetches, resulting in non-coalesced access that wastes $\sim$50\% bandwidth. Consequently, color attribute loading becomes the system bottleneck.

To address the above challenges, we present \we, an algorithm-system co-design for efficient training in 3DGS. 

\textbf{\textit{Technique-1}: \underline{Heuristic workload-sensitive} Gaussian density control.}
Current Gaussian splatting suffers from severe density imbalance (100× deviation). We find 80\% redundancy in dense regions while sparse areas need just 20\% more Gaussians. At the algorithm level, our solution dynamically adjusts density via statistical merging and compensated densification, reducing deviation by 49\% without quality loss. 

\textbf{\textit{Technique-2}: \underline{Similarity-based} Gaussian sampling and merging.}
Standard thread-to-pixel mapping creates severe bottlenecks (40× workload gap). We find 68\% computation wastes on redundant Gaussians, with 85\% mergeable in dense regions. At the system level, our solution combines color-hash sampling and adaptive thread merging, achieving 1.33$\times$ speedup with negligible quality loss.

\textbf{\textit{Technique-3}: \underline{Reordering-based} memory access mapping strategy.}
Baseline RGB storage causes non-coalesced memory access. At the mapping level, we restructure data into SOA format $[R_1,G_1,B_1...R_n,G_n,B_n]$, enabling warp-level batch loading with shared memory buffering. This achieves 92\% coalesced reads and 1.40$\times$ memory access speedup.

Experimental results demonstrate that \we achieves 1.44$\times$ speedup on end-to-end training compared with the  baseline 3DGS. Our work establishes that optimal 3DGS performance requires co-optimizing three efficiency metrics simultaneously: quality per Gaussian (geometric), processing speed per pixel (computational), and bytes per transaction (memory).

\section{Background and Related Work}

\subsection{3D Gaussian Splatting}

3D Gaussian Splatting (3DGS)~\cite{kerbl3Dgaussians} represents a significant advancement in explicit 3D scene representation, offering a compelling alternative to traditional implicit neural representations~\cite{mildenhall2021nerf}. 
3DGS employs a collection of anisotropic 3D Gaussian ellipsoids as its fundamental building blocks.

The core of 3DGS lies in its parameterization of scene elements as Gaussian primitives. Each Gaussian is defined by its center $\mu \in \mathbb{R}^3$ and a covariance matrix $\Sigma \in \mathbb{R}^{3 \times 3}$ that determines its spatial extent and orientation. 
The influence of each Gaussian at any point $x$ in space follows:

$$
G(x) = e^{-\frac{1}{2} (x - \mu)^\top \Sigma^{-1} (x - \mu)},
$$

The rendering pipeline of 3DGS involves several key steps. First, 3D Gaussians are projected onto the 2D image plane through a process called splatting. The projection transforms the 3D covariance $\Sigma$ to its 2D counterpart $\Sigma'$ using:

$$
\Sigma' = J W \Sigma W^\top J^\top,
$$

where $W$ represents the viewing transformation matrix and $J$ is the projective transformation. 
The final pixel colors are computed through front-to-back alpha blending:

$$
C = \sum_{i \in N} c_i \alpha_i \prod_{j=1}^{i-1} (1 - \alpha_j),
$$

where $N$ denotes the set of Gaussians contributing to the pixel, sorted by depth.

\subsection{Acceleration of 3D Gaussian Splatting}

Recent efforts \cite{bagdasarian20243dgszipsurvey3dgaussian} to accelerate 3D Gaussian Splatting (3DGS) focus on reducing the number of Gaussians, memory compression~\cite{lu2024scaffold} and multi-GPU parallelism~\cite{zhao2024scaling3dgaussiansplatting, chen2024dogsdistributedorientedgaussiansplatting}. EAGLES \cite{girish2023eagles} introduces a mechanism to control the growth of Gaussians by leveraging convergence curves, ensuring the process remains computationally efficient, while GES \cite{hamdi2024ges} employs a learning-based approach aimed at limiting the creation of high-frequency Gaussians, thereby optimizing the rendering pipeline. LightGaussian \cite{fan2023lightgaussian} takes a pruning-based approach, removing Gaussians based on their contribution to the volume, opacity, and pixel impact. C3DGS \cite{lee2024compact} uses scaling and opacity for pruning, integrating a masking loss to balance quality and speed. C3DGS \cite{lee2024compact} utilizes scaling and opacity for Gaussian pruning while incorporating a masking loss function to strike a balance between rendering quality and computational speed.

\section{Heuristic Workload-sensitive Density Control}
\begin{figure}
    \centering
    \includegraphics[width=0.99\linewidth]{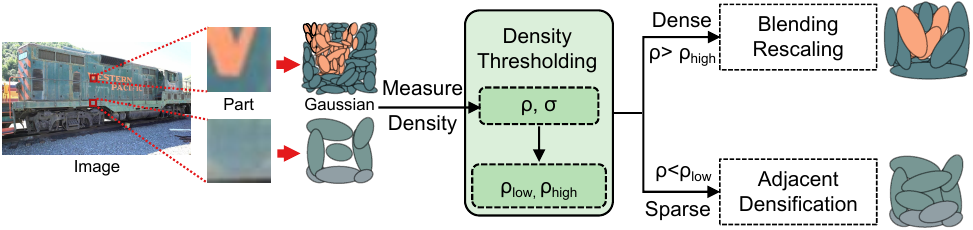}
    \vspace{-18pt}
    \caption{The pipeline of heuristic workload-sensitive density control. We first evaluate Gaussian density and determine the density threshold. Then, we perform adjacent densification and apply blending and rescaling based on the respective density magnitudes.}
    \vspace{-15pt}
    \label{fig:Approach1}
\end{figure}
\subsection{Motivation and Challenges}
Current 3D Gaussian Splatting methods \cite{kerbl3Dgaussians} face a fundamental density-quality dilemma. The adaptive densification strategy creates severe imbalances - high-gradient regions accumulate redundant Gaussians (up to 100$\times$ denser than sparse areas) while low-texture regions suffer from insufficient coverage. The core challenge lies in \textbf{skewed density allocation in dense and sparse regions during Gaussian densification}. This results in $\sim$80\% Gaussian waste in dense regions due to premature termination, while sparse regions exhibit artifacts from insufficient coverage.

\subsection{Analysis and Insights}
Through systematic analysis, we uncover key insights about this density-quality tradeoff. In dense regions, Reducing Gaussian counts by 20\% preserves accuracy while lowering computational costs by 35\%, revealing significant redundancy in current approaches. Conversely, sparse regions demonstrate that increasing density by just 20\% significantly eliminates artifacts, proving that current undersampling substantially harms quality. Furthermore, we observe that fixed density criteria fundamentally fail to account for evolving point distributions during optimization. These findings collectively reveal that \textbf{adaptive density regulation} must simultaneously consider both local geometric complexity and rendering workload distribution to achieve optimal efficiency.

\subsection{Approach}
The proposed heuristic workload-sensitive Gaussian density control method employs a statistically driven framework to dynamically manage Gaussian point density. As is shown in Fig. \ref{fig:Approach1}, the key components of this approach include adaptive thresholding, statistically guided densification, and merging.
\subsubsection{Statistical Density Thresholding}
Instead of manually defining density thresholds, we calculate them dynamically based on global density statistics. For each point \( p \), the local density \( \rho \) is determined as the number of neighboring points within a fixed radius \( r \). The density thresholds \( \rho_{\text{low}} \) and \( \rho_{\text{high}} \) are defined as:
\[
\rho_{\text{low}} = \mu_\rho - \alpha \cdot \sigma_\rho, \quad \rho_{\text{high}} = \mu_\rho + \beta \cdot \sigma_\rho,
\]
where \( \mu_\rho \) and \( \sigma_\rho \) are the mean and standard deviation of all local densities, and \( \alpha \) and \( \beta \) are fixed scaling factors (e.g., \( \alpha = 1, \beta = 1 \)). These thresholds adapt to the global distribution of points, ensuring scene-wide applicability without manual tuning.

\subsubsection{Adjacent Densification with Perturbation}

To increase density in low-density regions (\( \rho < \rho_{\text{low}} \)), new points are generated in adjacent areas using a Gaussian-based method. For each existing point \( p \), the average distance to its \( k \)-nearest neighbors is computed as:
\[
\bar{d}_p = \frac{1}{k} \sum_{i=1}^k d(p, p_i),
\]
where \( d(p, p_i) \) represents the Euclidean distance between \( p \) and its \( i \)-th neighbor. The standard deviation for the Gaussian distribution is then defined as:
\[
\sigma_p = \alpha \cdot \bar{d}_p, \quad \alpha > 1,
\]
ensuring that the new points are dispersed into adjacent areas. Locations for new points are sampled from a Gaussian distribution:
\[
p_{\text{new}} \sim \mathcal{N}(p, \sigma_p^2).
\]
To further introduce variability and avoid clustering artifacts, each sampled point is perturbed by a small random factor:
\[
p_{\text{new}} = p_{\text{new}} + \epsilon, \quad \epsilon \sim \mathcal{U}(-\delta, \delta).
\]
This process is repeated iteratively until the desired density is achieved, ensuring that new points are distributed near existing ones without overlap.

\subsubsection{Statistical Merging and Rescaling}
In high-density regions (\( \rho > \rho_{\text{high}} \)), redundant points are merged to reduce computational overhead. Pairs of points within a merging radius \( d_{\text{merge}} \) are identified, where \( d_{\text{merge}} \) is defined dynamically as:
\[
d_{\text{merge}} = \mu_d + \gamma \cdot \sigma_d,
\]
with \( \mu_d \) and \( \sigma_d \) representing the mean and standard deviation of distances between neighboring points, and \( \gamma \) being a scaling factor (e.g., \( \gamma = 1 \)). The merged point's location is computed as the opacity-weighted centroid:
\[
p_{\text{merged}} = \frac{\sum_i \omega_i p_i}{\sum_i \omega_i}, \quad \omega_i = \text{opacity}(p_i),
\]
and its scale is set to the mean scale of the original points. This approach preserves visual fidelity while optimizing computational efficiency.

\begin{figure}
    \centering
    \includegraphics[width=1\linewidth]{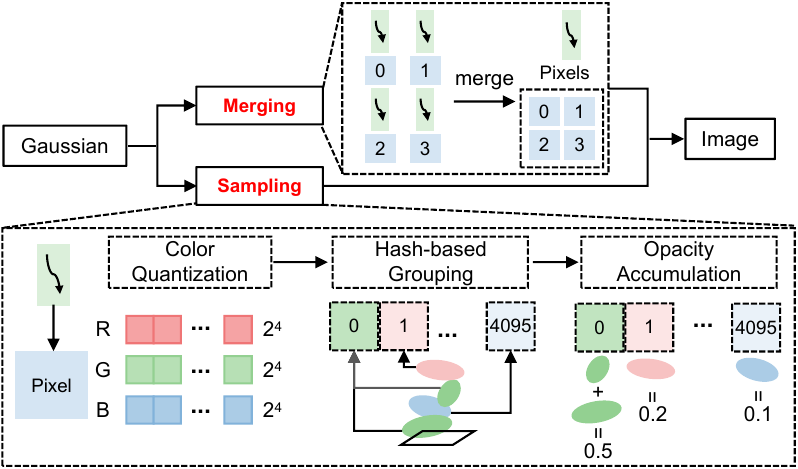}
    \vspace{-18pt}
    \caption{Similarity-based Gaussian sampling and merging. When merging, threads with light workload process four pixels. Similarity-based sampling includes color quantization, hash-based grouping and opacity accumulation.}
    \vspace{-15pt}
    \label{fig:Approach2}
\end{figure}
\subsubsection{Iterative Adaptation}
After specified iterations (1500 iterations in 3DGS), global density statistics are recalculated to update \( \rho_{\text{low}} \), \( \rho_{\text{high}} \), and \( d_{\text{merge}} \). This iterative process ensures that density control adapts dynamically to changes in point distributions, maintaining balanced workloads across the scene.


\section{Similarity-based Gaussian Sampling and Merging}
\subsection{Motivation and Challenge}

The standard one-to-one allocation strategy between threads and pixels of backend operator kernel creates severe processing bottlenecks in 3D Gaussian Splatting. Our measurements reveal that threads handling dense Gaussian clusters bear 40$\times$ more workload than those processing sparse regions (Fig.~\ref{fig:overview}-C2), while some threads remain idle during critical rendering phases. This imbalance stems from rigid thread scheduling that ignores spatial variations in Gaussian complexity, resulting in suboptimal GPU utilization and approximately 20\% training latency overhead. 
The core challenge is \textbf{imbalanced computation workload during Gaussian projection}. Such 40$\times$ workload gap forces finished threads to wait for unfinished ones.

\subsection{Analysis and Insight}

Our analysis highlights that the workload imbalance and dataflow inefficiency stem from the non-uniform distribution and varying importance of Gaussians. Many of these Gaussians show high color similarity, indicating that they can be computed fewer times, especially when viewed from certain perspectives.
Approximately 68\% of computational resources are wasted on processing visually redundant Gaussians in dense clusters. Second, the color similarity distribution shows that 85\% of neighboring Gaussians in dense regions can be merged with least quality impact. These findings lead to our crucial insight: \textbf{precise evaluation of each Gaussian is unnecessary when many share similar visual properties}.

\subsection{Approach}

To tackle the identified inefficiencies, we propose \textbf{similarity-based Gaussian sampling and merging}, a combined strategy comprising similarity-based sampling with color hashing, adaptive thread-block merging, and efficient data flow optimization for training and rendering, shown in Fig. \ref{fig:Approach2}.

\subsubsection{\textbf{Similarity-based Sampling with Color Hashing}}
Instead of a fine-grained approach, which incurs high memory overhead, we utilize a coarse-grained method that reduces memory usage and computational costs. Specifically, we group $16 \times 16$ pixel blocks that share the same set of Gaussians, minimizing the memory footprint. This block size was determined empirically to balance between memory efficiency (favoring larger blocks) and color accuracy (favoring smaller blocks). To further optimize the color similarity evaluation, we employ a color hashing technique based on quantized RGB values.

\paragraph{Color Quantization}
We discretize the RGB color space by dividing each color channel (R, G, B) into $2^4$ levels, resulting in $2^{12} = 4096$ distinct color buckets. The quantization process uses uniform binning with the following mapping function for each channel $c$:
$$
c_{\text{quant}} = \left\lfloor \frac{c}{256} \times 16 \right\rfloor
$$
where $c$ is the 8-bit color value (0-255). Each pixel is mapped to one of these discrete color buckets based on its quantized RGB values. 

\paragraph{Hash-based Grouping}
During the traversal of the data, each pixel is assigned to a corresponding color bucket based on its quantized color value. We implement this using a hashing scheme where the hash function is simply:
$$
\text{hash}(R_q, G_q, B_q) = R_q \times 256 + G_q \times 16 + B_q
$$
where $R_q, G_q, B_q$ are the quantized color values. The hash table is implemented using shared memory on the GPU. Pixels within the same bucket are considered similar in color, allowing us to directly merge their opacity values without detailed pairwise similarity computation.

\paragraph{Bucket-wise Aggregation}
For each color bucket, we aggregate the gaussians by averaging the color and accumulating the opacity. The aggregation process uses a two-stage reduction: first within thread blocks using shared memory, then across blocks using global memory. This eliminates the need for fine-grained comparisons, reducing both memory usage and computational overhead. The result is a set of representative colors and their associated opacity values for each bucket. We maintain a counter for each bucket to track the number of contributing pixels, enabling proper weighted averaging of colors. 

This hash-based approach reduces the time complexity of the color similarity evaluation from $O(n^2)$ to $O(n)$, as each data point is processed once during quantization and once during aggregation. 

\subsubsection{\textbf{Adaptive Thread-Block Merging}}
To address the workload imbalance, we implement an adaptive thread-block merging strategy. The adaptation is based on a density metric $\rho$ computed for each region:
$$
\rho = \frac{\text{number of active Gaussians}}{\text{number of pixels}}
$$
In semi-sparse regions, each thread processes two pixels, while in sparse regions, it handles four pixels, reducing thread management overhead. For dense regions, we maintain the standard one-pixel-per-thread assignment to preserve quality. This merging reduces redundant computations and improves resource utilization. 

\subsubsection{\textbf{Efficient Data Flow Optimization for Training and Rendering}}
Building on the coarse-grained sampling and merging framework, we integrate data flow optimization to further enhance efficiency during both training and rendering phases. We introduce a two-level caching hierarchy: During training, we perform an importance analysis of Gaussians for each viewpoint, leveraging a color similarity metric derived from the sampling process. The importance score $I_g$ for Gaussian $g$ is computed as:
$$
I_g = \frac{1}{N}\sum_{i=1}^N \text{similarity}(c_g, c_i) \times \alpha_i
$$
where $c_g$ is the Gaussian's color, $c_i$ are neighboring pixel colors, and $\alpha_i$ are their opacities. Gaussians with distinct colors are prioritized, while those with high color similarity are deemed less significant. 

Incorporating this analysis into the training process enables a focused allocation of computational resources to the most impactful Gaussians, effectively reducing training time without compromising image quality. Additionally, the precomputed importance information is retained for use during rendering.  This selective evaluation eliminates unnecessary computations, significantly accelerating the image generation process, particularly in real-time applications.

\section{Reordering-based Memory Access Mapping Strategy}
\subsection{Motivation and Challenge}\label{tech3_challenge}
The baseline 3DGS method demonstrates inefficiencies in memory access due to the way RGB color (feature) data is organized. The RGB data is stored in a non-contiguous format, where all red values are stored consecutively, followed by green and blue values. This arrangement results in non-coalesced memory access patterns during rendering, causing delays in processing.

Profiling results reveal that accessing RGB data from global memory incurs significant latency, with each access requiring approximately 200-300 ns. This inefficiency doubles rendering time compared to an ideal scenario with optimal memory access. Additionally, global memory reads for RGB values account for almost 40\% of the total rendering time. The scattered storage format forces each thread to retrieve data from disparate memory locations, leading to under-utilization of GPU resources and performance loss.

The challenge is \textbf{the fragmented memory access during color splatting, resulting in low bandwidth utilization and training efficiency.}
Optimizing the retrieval mechanism for RGB data is essential to achieve faster rendering and improve overall system performance.
\subsection{Analysis and Insight}
According to profiling results in Sec.\ref{tech3_challenge}, it becomes evident that \textbf{aligning memory access patterns} with GPU hardware preferences can enhance performance. Additionally, \textbf{leveraging shared memory for intermediate data storage} offers the potential to reduce the overhead associated with global memory reads. 

\begin{figure}[!t]
    \centering
    \includegraphics[width=0.99\linewidth]{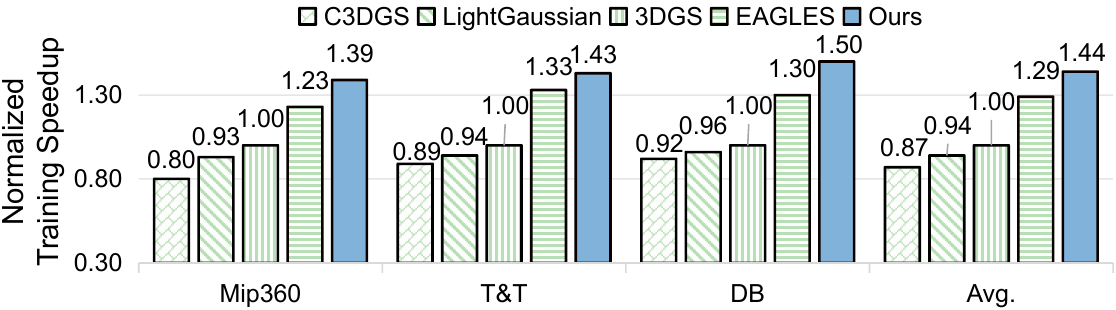}
    \vspace{-22pt}
    \caption{Evalutation of training speedup on Mip-NeRF360, Tanks\&Temples, and Deep Blending datasets.}
    \vspace{-5pt}
    \label{fig:stat}
\end{figure}

\begin{table}[!t]
\caption{Evaluation of synthesis quality}
\vspace{-10pt}
\centering
\resizebox{\columnwidth}{!}{%
\begin{tabular}{c|ccc|ccc|ccc}
\toprule
\textbf{Datasets} & \multicolumn{3}{c|}{\textbf{MipNeRF-360}} & \multicolumn{3}{c|}{\textbf{Tank\&Temples}} & \multicolumn{3}{c}{\textbf{Deep Blending}} \\ 
\cmidrule(lr){2-4} \cmidrule(lr){5-7} \cmidrule(lr){8-10}
\textbf{Method}        & \textbf{SSIM↑}     & \textbf{PSNR↑}    & \textbf{LPIPS↓}    & \textbf{SSIM↑}     & \textbf{PSNR↑}     & \textbf{LPIPS↓}     & \textbf{SSIM↑}     & \textbf{PSNR↑}     & \textbf{LPIPS↓} \\ 
\midrule
3DGS          & 0.798     & 28.95    & 0.193     & 0.845     & 23.52     & 0.360      & 0.901     & 29.58     & 0.242      \\
LightGaussian & 0.858     & 28.63    & 0.213     & 0.816     & 23.06     & 0.232      & 0.876     & 28.39     & 0.304      \\
C3DGS         & 0.747     & 28.01    & 0.219     & 0.823     & 22.98     & 0.432      & 0.881     & 28.67     & 0.293      \\
EAGLES        & 0.736     & 28.58    & 0.202     & 0.813     & 22.78     & 0.478      & 0.836     & 28.98     & 0.301      \\
\midrule
Ours          & 0.756     & 28.03    & 0.217     & 0.801     & 22.82     & 0.401      & 0.848     & 28.81     & 0.278      \\
\bottomrule
\end{tabular}%
}
\label{tab:metrics}
\vspace{-15pt}
\end{table}

\begin{table}[!t]
\caption{Ablation Study on synthesis quality}
\vspace{-10pt}
\centering
\resizebox{\columnwidth}{!}{%
\begin{tabular}{c|ccc|ccc|ccc}
\toprule
\textbf{Datasets} & \multicolumn{3}{c|}{\textbf{MipNeRF-360}} & \multicolumn{3}{c|}{\textbf{Tank\&Temples}} & \multicolumn{3}{c}{\textbf{Deep Blending}} \\ 
\cmidrule(lr){2-4} \cmidrule(lr){5-7} \cmidrule(lr){8-10}
\textbf{Method} & \textbf{SSIM↑} & \textbf{PSNR↑} & \textbf{LPIPS↓} & \textbf{SSIM↑} & \textbf{PSNR↑} & \textbf{LPIPS↓} & \textbf{SSIM↑} & \textbf{PSNR↑} & \textbf{LPIPS↓} \\ 
\midrule
3DGS      & 0.798  & 28.95  & 0.193  & 0.845  & 23.52  & 0.360  & 0.901  & 29.57  & 0.242  \\
3DGS + T1 & 0.785  & 28.78  & 0.202  & 0.838  & 23.36  & 0.375  & 0.882  & 29.27  & 0.252  \\
3DGS + T2 & 0.779  & 28.42  & 0.208  & 0.823  & 23.09  & 0.389  & 0.857  & 29.03  & 0.269  \\
3DGS + T3 & 0.798  & 28.94  & 0.193  & 0.845  & 23.52  & 0.360  & 0.901  & 29.57  & 0.242  \\
\midrule
Ours      & 0.756  & 28.03  & 0.217  & 0.801  & 22.82  & 0.401  & 0.848  & 28.81  & 0.278  \\
\bottomrule
\end{tabular}%
}
\label{abltiontable}
\vspace{-10pt}
\end{table}

\vspace{-5pt}
\begin{figure*}[!t]
    \centering
    \includegraphics[width=0.99\linewidth]{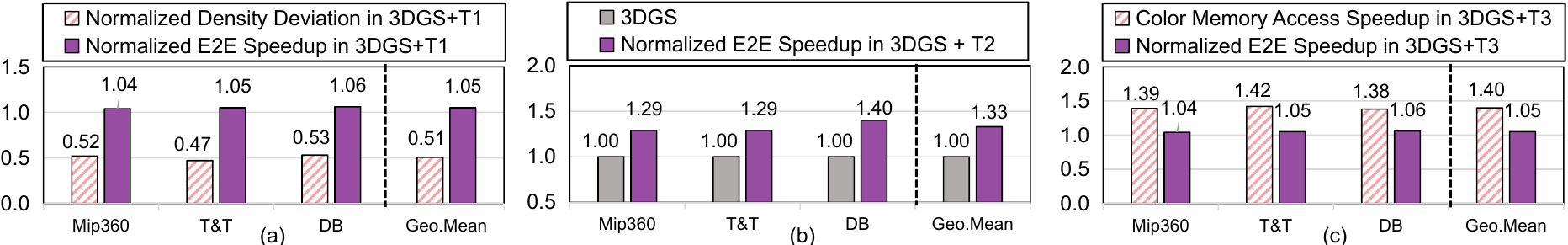}
    \vspace{-10pt}
    \caption{Ablation study. (a) Normalized density deviation and end-to-end speedup with workload-sensitive density control (T1). (b) Normalized end-to-end speedup with similarity-based sampling and merging (T2). (c) Color memory access speedup and normalized end-to-end speedup with reordering-based consecutive color memory access (T3).}
    \vspace{-18pt}
    \label{ablation2}
\end{figure*}

\subsection{Approach}

To address the identified inefficiencies, we propose reordering-based memory access mapping strategy, a three-step optimization comprising batch loading, data reordering, and coalesced memory access. As shown in Fig.~\ref{fig:overview}, this approach leverages shared memory to improve memory operation efficiency and accelerate rendering performance.

The first step, \textbf{batch loading}, preloads RGB color data into shared memory in batches rather than retrieving it from global memory during each rendering step. Shared memory, with its lower latency and higher bandwidth compared to global memory, significantly reduces the frequency and cost of global memory accesses, thereby minimizing delays caused by repeated data retrievals.

The second step, \textbf{data reordering}, reorganizes the RGB data format in shared memory. In the baseline method, all red values were stored first, followed by green and blue values, resulting in a scattered layout. This step transforms the format into a contiguous layout, storing the RGB values for each Gaussian point together. By aligning the data layout with the GPU's memory access patterns, this reordering enables more efficient data retrieval.

The final step, \textbf{coalesced memory access}, optimizes memory retrieval during rendering. The reordered data in shared memory allows GPU threads to access RGB values for each Gaussian point in a coalesced manner. This streamlined access pattern capitalizes on shared memory's fast caching capabilities, reducing the time spent on memory reads and enhancing overall throughput.

By combining these three steps, our reordering-based memory access mapping strategy  reduces dependence on slower global memory and aligns memory access patterns with GPU architecture.

\vspace{-5pt}

\section{Experiments}\label{AA}

\subsection{Models and Datasets}
To validate the effectiveness of our method, we utilized the same datasets and evaluation metrics as 3DGS~\cite{kerbl3Dgaussians}. 
Specifically, the datasets include Mip-NeRF360 (abbreviated as Mip360)~\cite{barron2021mip}, Tanks\&Temples (abbreviated as T\&T)~\cite{Knapitsch2017}, and Deep Blending (abbreviated as DB)~\cite{DeepBlending2018}, encompassing both indoor and outdoor environments. 
It is important to note that the undocumented flower and treehill scenes were excluded from the evaluation. 
Performance metrics including training time, model size, and synthetic quality, are evaluated using PSNR, SSIM~\cite{1284395}, and LPIPS~\cite{zhang2018unreasonableeffectivenessdeepfeatures}. 
All experiments are conducted on a single NVIDIA A100 GPU.

\subsection{Baselines}
We evaluate the proposed \we framework against 3DGS~\cite{kerbl3Dgaussians} and state-of-the-art accelerated 3DGS rendering methods, including LightGaussian~\cite{fan2023lightgaussian}, EAGLES~\cite{girish2023eagles}, and C3DGS~\cite{Lee_2024_CVPR}. We focus on the training speedup and synthesis quality between \we and baselines.

\subsection{Evaluation of Synthesis Quality}
As illustrated in Table \ref{tab:metrics}, we compare the synthesis quality with four existing methods. 
Taking PSNR for MipNeRF-360 as an example, compared to 3DGS~\cite{kerbl3Dgaussians}, \we slightly reduces the synthesis quality from 28.95 to 28.03.
Compared to LightGaussian~\cite{fan2023lightgaussian}, \we slightly reduces the synthesis quality from 28.63 to 28.03.
Compared to C3DGS~\cite{Lee_2024_CVPR}, \we improves the synthesis quality from 28.01 to 28.03.
Compared to EAGLES~\cite{girish2023eagles}, \we slightly reduces the synthesis quality from 28.58 to 28.03.

\subsection{Evaluation of Training Speed}
As illustrated in Fig. \ref{fig:stat}, we benchmark the original 3DGS and various existing methods. Our \we framework demonstrated the fastest training times among all evaluated approaches. Specifically, compared to 3DGS~\cite{kerbl3Dgaussians}, \we reduces training time by an average of 1.44$\times$ while preserving synthesis quality comparable to the baseline.
compared to EAGLES~\cite{girish2023eagles}, \we reduces training time by an average of 1.12$\times$ and achieves better synthesis quality.

\vspace{-2pt}
\subsection{Ablation Study}

Beginning with the baseline 3DGS framework, we integrate each technique: \textit{Heuristic Workload-sensitive Density Control}, \textit{Similarity-based Gaussian Sampling and Merging}, and \textit{Reordering-based Memory Access Strategy}, presented in Fig.~\ref{ablation2} and Table \ref{abltiontable}.

\subsubsection{Heuristic Workload-sensitive Density Control}
As shown in Fig.~\ref{ablation2}(a), experimental results demonstrate that our method reduces density deviation to 51\% of its original value.
It also achieves 1.05$\times$ training speedup with minimal impact on rendering quality. 

\subsubsection{Similarity-based Gaussian Sampling and Merging}
As shown in Fig.~\ref{ablation2}(b), our method achieves 1.33$\times$ training speedup while incurring negligible loss in rendering quality. This acceleration is achieved by \textbf{eliminating redundant computations} through the reduction of unnecessary Gaussian sphere calculations.

\subsubsection{Reordering-based Memory Access Strategy}
As shown in Fig.~\ref{ablation2}(c), our method enhances the speed of color memory access by 1.40$\times$. 
And it achieves 1.06$\times$ training speedup without compromising rendering quality.

\vspace{-7pt}
\section{Conclusion}
We present \we, a novel framework designed to optimize 3D Gaussian Splatting (3DGS) on GPUs. \we proposes heuristic workload-sensitive density control, similarity-based Gaussian sampling and merging, and reordering-based memory access mapping strategy. Experimental results demonstrate that \we achieves 1.44$\times$ training speedup than baseline 3DGS while maintaining high rendering quality. 

\section*{Acknowledgment}
This work was sponsored by Shanghai Rising-Star
Program (No. 24QB2706200) and Beijing Douyin Information Service Co., Ltd.

\bibliographystyle{unsrt}
\bibliography{ref}

\end{document}